\documentclass{article}
\usepackage{ijcai22}

\usepackage{times}
\usepackage{soul}
\usepackage{url}
\usepackage[hidelinks]{hyperref}
\usepackage[utf8]{inputenc}
\usepackage[small]{caption}
\usepackage{graphicx}
\usepackage{amsmath}
\usepackage{amsthm}
\usepackage{amssymb}  
\usepackage{amsfonts} 
\usepackage[labelformat=empty]{subcaption} 
\usepackage{booktabs}
\usepackage{algorithm}
\usepackage{algorithmic}
\usepackage{lipsum}

\title{End-to-End Policy Gradient Method for POMDPs and Explainable Agents}

\author{
Soichiro Nishimori$^{1}$\footnote{This work is done at Kyoto University.}
\and
Sotetsu Koyamada$^{2, 3}$\And
Shin Ishii$^{2, 3, 4}$
\affiliations
$^1$ Graduate School of Frontier Sciences, The University of Tokyo\\
$^2$ Graduate School of Informatics, Kyoto University, Kyoto, Japan\\
$^3$ Advanced Telecommunications Research Institute International (ATR), Kyoto, Japan\\
$^4$ International Research Center for Neurointelligence, The University of Tokyo, Tokyo, Japan
\emails
nishimori-soichiro358@g.ecc.u-tokyo.ac.jp,
koyamada-s@sys.i.kyoto-u.ac.jp,
ishii@i.kyoto-u.ac.jp
}

\begin{document}

\maketitle

\begin{abstract}
Real-world decision-making problems are often partially observable, and many can be formulated as a Partially Observable Markov Decision Process (POMDP). 
When we apply reinforcement learning (RL) algorithms to the POMDP, reasonable estimation of the hidden states can help solve the problems.
Furthermore, explainable decision-making is preferable, considering their application to real-world tasks such as autonomous driving cars.
We proposed an RL algorithm that estimates the hidden states by end-to-end training,  and visualize the estimation as a state-transition graph.
Experimental results demonstrated that the proposed algorithm can solve simple POMDP problems and that the visualization makes the agent's behavior interpretable to humans.
\end{abstract}

\section{Introduction}

Reinforcement learning (RL) is a framework for solving optimal control problems in a Markov Decision Process (MDP)~\cite{sutton2018reinforcement}. 
With the rapid increase in computational resources as background, deep RL, which utilizes multilayer neural networks as function approximators, has succeeded in complete-information games with huge search space, such as Go~\cite{silver2016mastering}. On the other hand, in reality, the MDP cannot formulate many optimal control problems. 

A POMDP is a decision-making problem where only a part of the true state is observable (i.e., the rest is not observable). 
the POMDP is challenging to solve compared with the MDP. 
In the MDP, the optimal policy is deterministic in general. In contrast, in the POMDP, the optimal policy is memory-based~\cite{monahan1982state}, and the best policy is stochastic when restricted to non-memory policy~\cite{singh1994learning}. 
However, POMDPs are an essential class of problems because they formulate many real-world problems
For example, in automated driving, the driver (agent) cannot access observations like a bird's eye view.
 Therefore, it is necessary to develop artificial intelligence (AI) agents capable of solving POMDPs.

Recently, the momentum for real-world applications of AI technology has been growing, and machine learning models that non-experts can interpret, i.e., explainable AI, are required~\cite{adadi2018peeking,gunning2019xai}. In this context, trained agents in POMDPs often lack explainability. In MDPs, the optimal policy is a deterministic function, while in POMDPs, the optimal policy can be more complex than a simple deterministic function.
It is difficult for humans to interpret a trained agent's behavior, even if optimal.

The contributions of this study are the following two points.
\begin{itemize}
\item We derived an end-to-end policy gradient-based RL algorithm for POMDPs by extending a prior work~\cite{Aberdeen02scalinginternal-state}. We demonstrated that it could solve simple POMDP problems only with small minor model changes to existing policy-gradient methods.

\item We proposed visualizing the discrete internal states learned by the proposed method to make the agent's learned policies interpretable to humans. We demonstrated it in two POMDP environments.
\end{itemize}

\section{Preliminaries}
This section introduces the basic formulation of RL and the framework to solve it and briefly summarizes the existing method.
\subsection{Objective of RL}
We outline the purpose of RL in the MDP. The following tuple determines the MDP $(\mathcal{S}, \mathcal{A}, p_0, P, r)$, where $\mathcal{S}$ is a state space, $\mathcal{A}$ is an action space, $p_0$ is a distribution of initial states, $P$ is a distribution of state transitions, and $r$ is a reward function. The agent acts according to a policy $\pi_\theta(a|s)$ in this environment, where $\theta$ is a parameter. We denote the expectation operator over a trajectory induced by a policy $\pi_\theta$ as $\mathbb{E}_{\pi_\theta}$.
In this paper, the finite horizon is assumed. The goal of RL is to obtain a policy that maximizes the expectation of the cumulative sum of rewards represented as
\begin{equation}
\label{eqn:reinfrocement_learning}
    J(\theta) = \mathbb{E}_{\pi_\theta}[R_\tau],
\end{equation}

where $R_\tau = \sum_{t=0}^{T}r(s_t, a_t)$.

\subsection{Policy Gradient Method}
The policy gradient method is an algorithm that directly optimizes the parameterized policy using the policy gradient~\cite{sutton2000policy}.
REINFORCE~\cite{williams1992simple} provides the simplest form of policy gradient estimation, which uses the Monte-Carlo return to estimate the expected return in the following policy gradient.
\begin{equation}
\label{eqn:policy_gradient}
   \nabla_{\theta} J(\theta) = \mathbb{E}_{\pi_\theta}\biggl[\sum_{t=0}^{T}\nabla \log\pi_\theta(a_t|s_t)R_\tau \biggl].
\end{equation}
We use REINFORCE algorithm to demonstrate our proposed method.
However, note that our proposed method is applicable to other policy gradient methods.

\subsection{POMDP}
Under MDPs, agents are given the states: the necessary information to take optimal action. POMDPs deal with the generalization of this situation. A POMDP is a decision-making problem where an agent is given only a subset of the true state as an observation. A POMDP is defined by the following tuple $( \mathcal{S} , \mathcal{A} ,\mathcal{O}, P_o , p_0, P, r )$, where $\mathcal{O}$ is an observation set and $P_o(o | s)$ defines the observation distribution given a state.

\begin{figure}[tb]
  \begin{center}
    \includegraphics[width=0.6\hsize]{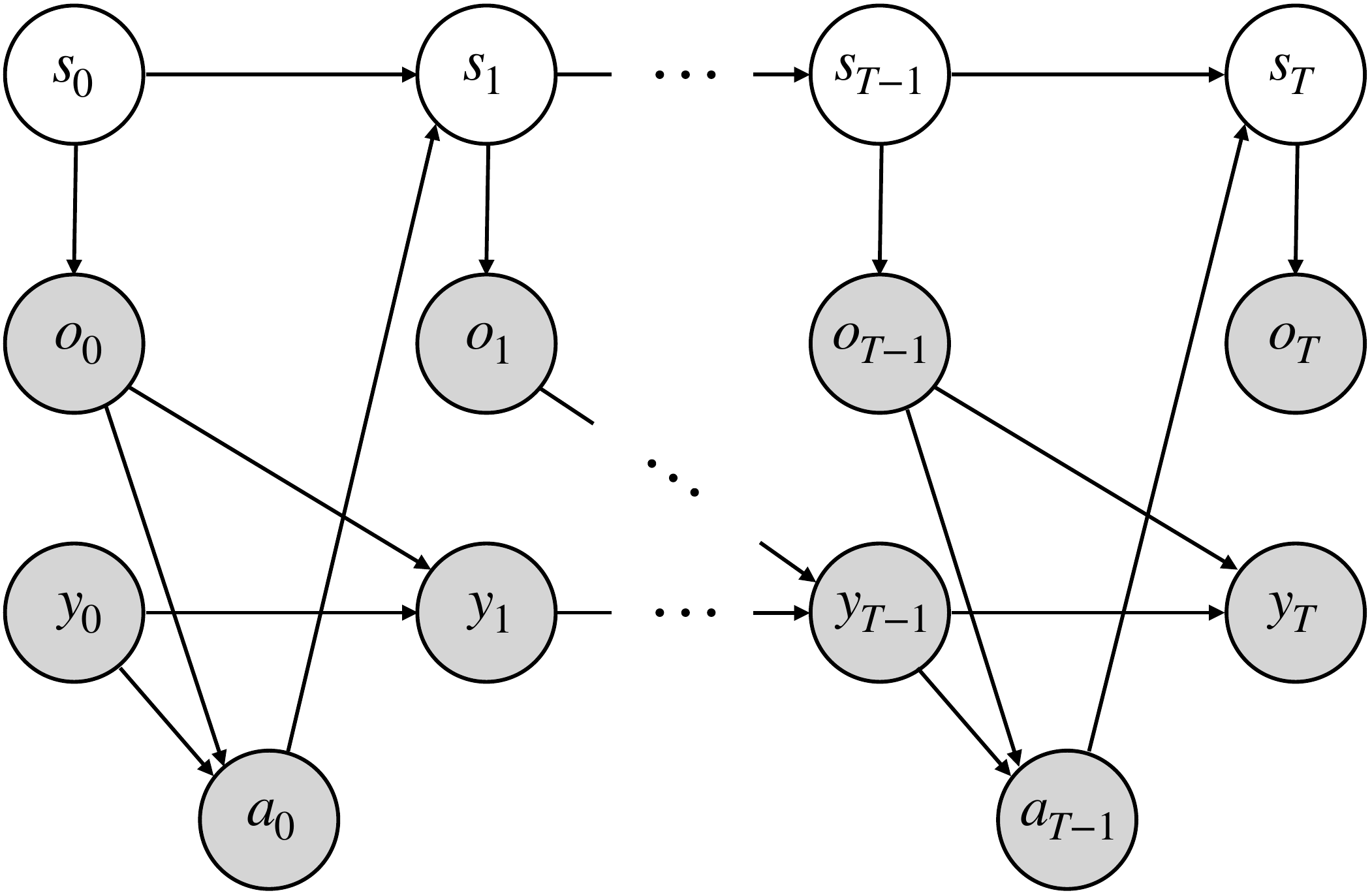}
    \caption{Graphical model with internal states.}
    \label{fig:graphical_model}
  \end{center}
\end{figure}

\subsection{Policy Gradient with Internal State}
We introduce a Markovian internal state $y \in \mathcal{Y}$ to solve POMDPs using the policy gradient method, where $\mathcal{Y}$ is the internal state space.
Then, we explain the existing method of estimating the policy gradient with the internal states. The graphical model is shown in Figure~\ref{fig:graphical_model}.
\subsubsection{Internal State}
Using internal states, we aim to construct an agent that behaves well under partial observation by estimating and retaining information necessary for decision-making that is not observable. We notate an internal state distribution as $\xi_\phi(y_t| o_{t-1}, y_{t-1})$, where $\phi$ is a parameter.
We represented a policy as a joint distribution of actions and internal states in this framework.

The joint policy is defined as follows
\begin{equation}
\label{eqn:joint_policy}
    \pi_{\theta,\phi}(a_t, y_t | o_t, o_{t-1}, y_{t-1}) = \pi_\theta(a_t| o_{t}, y_t)\xi_\phi(y_t| o_{t-1}, y_{t-1}).
\end{equation}
We assume internal states follow the categorical distribution and fix the initial internal state to zero.
The policy gradient with internal states is 
\begin{eqnarray}
\label{eqn:pgwi}
 && \nabla_{\theta, \phi} J(\theta, \phi) = \\
 && \,\,\,\,\,\, \mathbb{E}_{\pi_{\theta, \phi}} \biggl[\sum_{t=0}^{T}\nabla_{\theta, \phi} \log\pi_{\theta, \phi}(a_t, y_t|o_t, o_{t-1}, y_{t-1})R_\tau\biggl]. \nonumber 
\end{eqnarray}

\subsubsection{IState-GPOMDP}
IState-GPOMDP~\cite{Aberdeen02scalinginternal-state} decomposes the log term in the above policy gradient into the log of $\pi_{\theta}$ and $\xi_{\phi}$, and updates the parameters independently:
\begin{eqnarray}
 \nabla_{\theta} J(\theta) &=& \mathbb{E}_{\pi_{\theta, \phi}} \biggl[\sum_{t=0}^{T}\nabla_{\theta} \log\pi_\theta(a_t|o_t, y_t)R_\tau\biggl], \\
 \nabla_{\phi} J(\phi) &=& \mathbb{E}_{\pi_{\theta, \phi}} \biggl[\sum_{t=0}^{T}\nabla_{\phi} \log\xi_\phi(y_t|o_{t-1}, y_{t-1})R_\tau\biggl].
\end{eqnarray}
Each parameter can be optimized in the same way as usual policy gradient methods.

\begin{figure}[tb]
  \begin{center}
    \includegraphics[width=0.7\hsize]{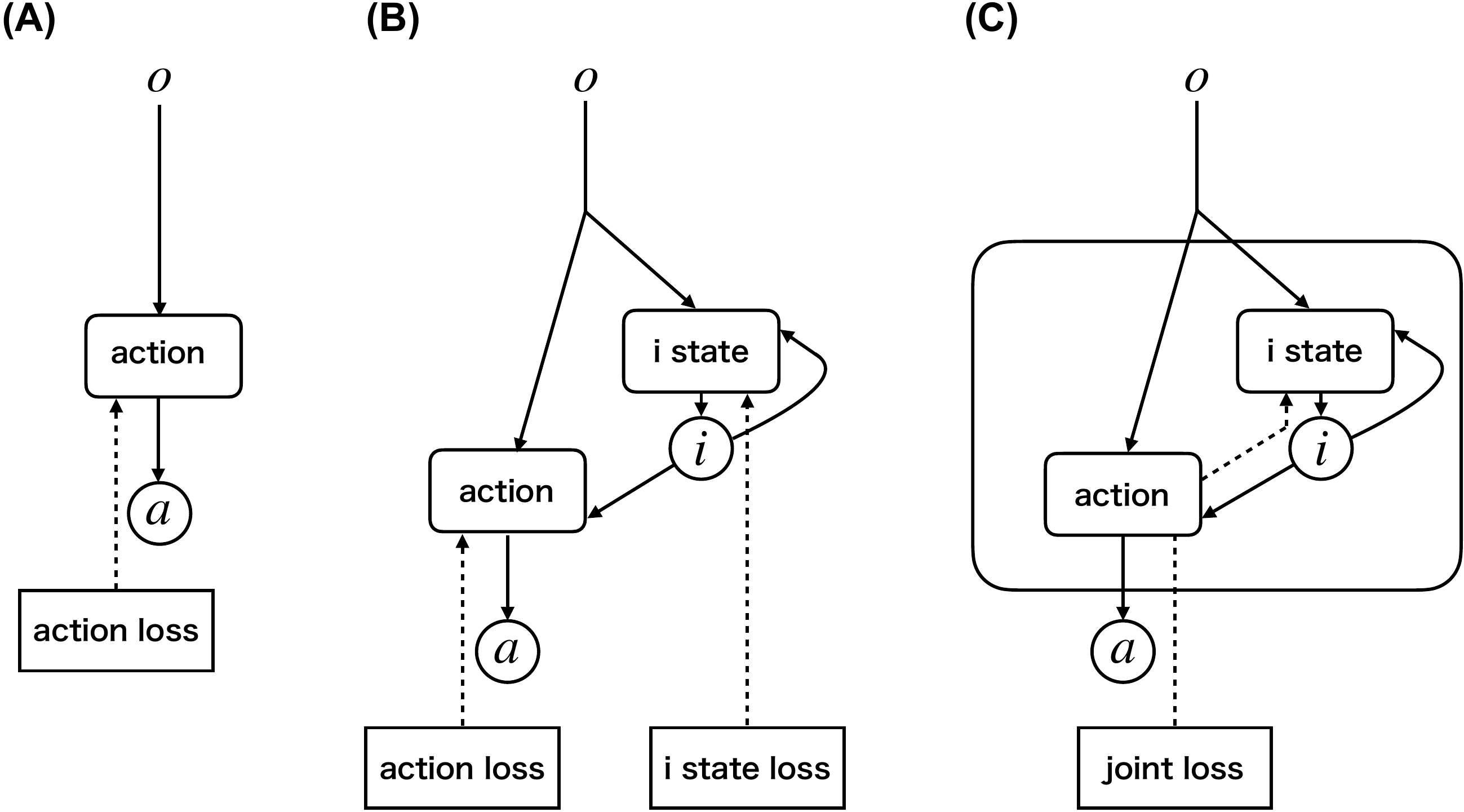}
    \caption{Gradient flow of each method. (A) normal policy gradient. (B) separate method (C) end-to-end. 
}
    \label{fig:gradient_flow}
  \end{center}
\end{figure}

\section{Method}
In this section, we describe how we train the joint policy end-to-end. Then, we explain how to create a transition graph to interpret the trained policy.
\subsection{End-to-End Architecture}
In deep learning literature, end-to-end learning refers to 
the method of training multiple neural network modules using backpropagation over the entire modules.
We propose to compute the policy gradient of Equation~\eqref{eqn:pgwi} by this end-to-end framework.
Figure~\ref{fig:gradient_flow} shows the comparison of the gradient computations.
The parameters of the internal state and the action models can be represented as a single parameter as follows
\begin{equation}
\label{eqn:end-to-end_policy}
\pi_{\psi}(a_t, y_t| o_t, o_{t-1}, y_{t-1}) = \pi_\theta(a_t| o_{t}, y_t)\xi_\phi(y_t| o_{t-1}, y_{t-1}),
\end{equation} 
where $\psi$ denotes the parameter of the networks in end-to-end learning.
The policy gradient is as follows
\begin{equation}
\label{eqn:end-to-end_pgwi}
\begin{split}
    \nabla_{\psi} J(\psi) = \mathbb{E}_{\pi_{\psi}} \biggl[\sum_{t=0}^{T}\nabla_\psi \log\pi_\psi(a_t ,y_t|o_t, o_{t-1}, y_{t-1})R_\tau\biggl].
\end{split}
\end{equation}
Compared with Equation~\eqref{eqn:policy_gradient}, it is clear that the end-to-end method allows us to train a joint policy in the same framework as the usual policy gradient method.

There are two benefits of this end-to-end training over the IState-GPOMDP, which we refer to as the separate method:
\begin{itemize}
    \item Just changing the neural network architecture is enough to apply the existing policy gradient method to POMDP problems. 
    \item The end-to-end method reduces the number of hyper-parameters compared to the separate method since it needs two separate neural networks.
\end{itemize}

\textbf{Implementation details.}
We can use the score function to approximate a gradient of function involving sampling operations. We can utilize this approximation in many deep-learning libraries, including PyTorch and TensorFlow. 
With this technique, we can represent a joint policy as a single neural network, which means end-to-end learning is possible in the framework. 

\begin{figure}[tb]
  \begin{center}
    \includegraphics[width=0.9\hsize]{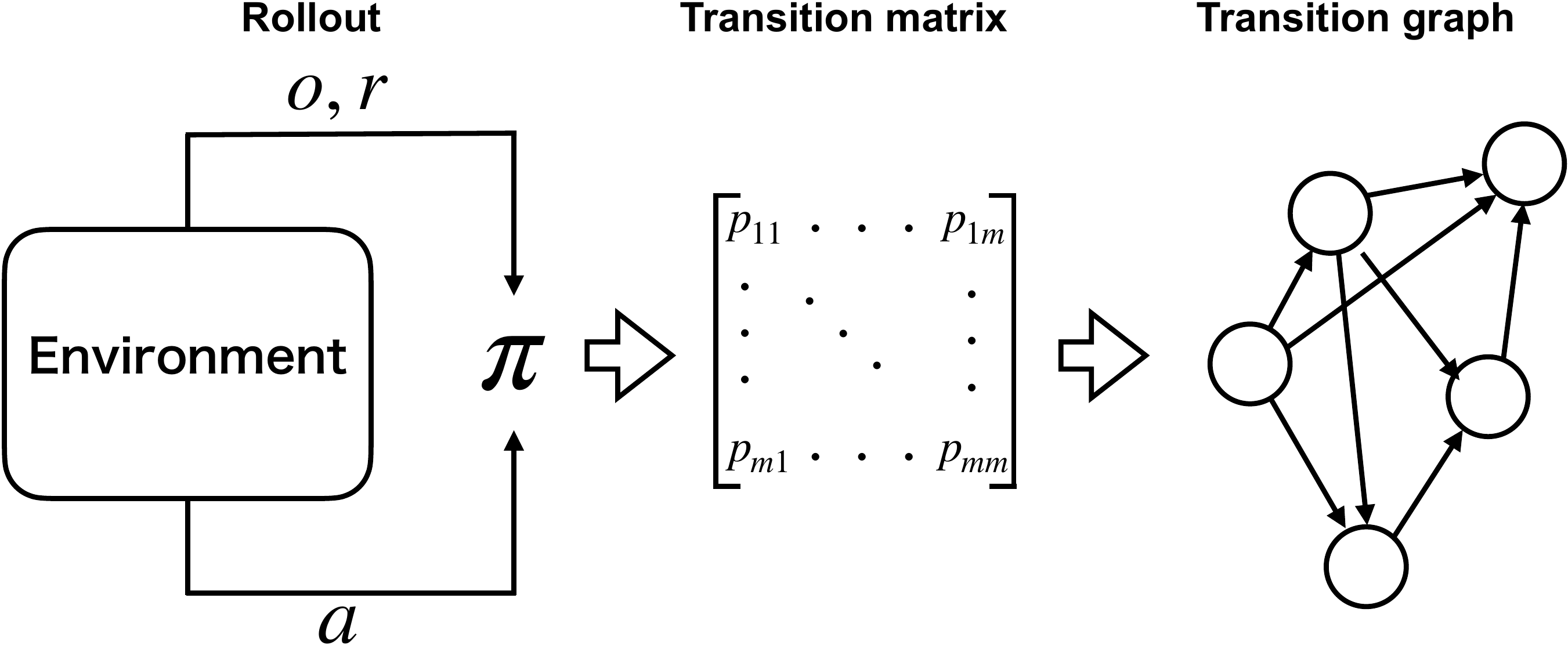}
    \caption{The process of creating a transition graph.}
    \label{fig:transition_graph_generation}
  \end{center}
\end{figure}

\subsection{Interpretability}
This study attempted to increase the interpretability of the agent's actions by visualizing the trained joint policy as the transition graph. 

Suppose an agent trained through the policy gradient method with internal states can successfully solve a POMDP problem. Then, we expect the internal states to store information necessary for solving the problem, which is not available from current observations.
Our goal is to provide that information to humans in an interpretable way.

We visualized the trained internal states by creating the transition graph of observation and internal state pairs. Creating a transition graph includes three procedures (shown in Figure~\ref{fig:transition_graph_generation}): (1) making a rollout using a trained model and gathering a sequence of actions, internal states, and observations. (2) calculating the pairs' empirical transition matrix from the collected data by marginalizing the actions. (3) dropping unreachable pairs from the matrix and creating a transition graph from the transition matrix.
The graph with a table of policy distribution over each pair of observations and internal states provides us with sufficient material for analyzing and interpreting the behavior of the trained model.

Our state-transition graph has similarities to the plan graph, where a policy is visualized in a graph manner~\cite{kaelbling1998planning}. The essential difference is that we visualize internal states by which we can interpret an agent's belief over hidden states.
\begin{figure}[htbp]
  \begin{center}
    \includegraphics[width=0.9\hsize]{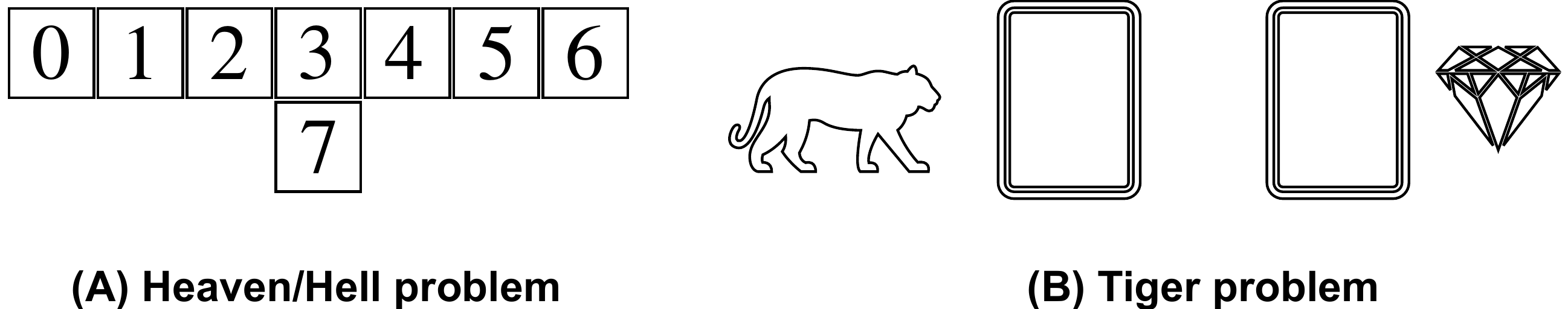}
  \caption{The illustration of the environments.}
  \label{fig:environment_illustration}
  \end{center}
\end{figure}

\section{Results}
We conducted experiments on two POMDP environments: the Heaven/Hell problem~\cite{bonet1998solving} and the Tiger problem~\cite{kaelbling1998planning}.
\subsection{Environments}
The Heaven/Hell problem is a problem in which an agent seeks to reach the goal based on the observation from the start point (the tile of 7 in Figure~\ref{fig:environment_illustration}). 
There are two goals in the Heaven/Hell problem: heaven, where the agent can receive a reward, and hell, where the agent receives a penalty. The reward and penalty are 1 and -1, respectively. The positions of heaven and hell are randomly determined from the two ends for each trial. 
This environment gives an agent a minor penalty every time, and the penalty is -0.01. The observation is a pair of the agent position and the heaven position. The agent position is always observable, while the heaven position is observable only in tile 7. In the other tiles, -1 is given as the heaven position. The actions are up, right, down, and left. In order to solve this problem, it is necessary to remember information about the position of heaven, at least up to the branching point.

The second problem is the Tiger problem. The Tiger problem is a task to open one of two doors. There are two doors, one with a tiger to receive a penalty and the other with a reward. The reward and the penalty are 0.1 and -1, respectively. The possible actions are right (open the door on the right), left (open the left door), and listen (listen to the sound behind the door in exchange for a minor penalty). The miner penalty for listening is -0.01. With probability $p$, the agent will be given the wrong information—the probability is 0.15. The observations are left, right, and null (initial signal). In the Tiger problem, repeatedly listening to the sound behind the door can provide correct information. An agent is required to memorize the location of the tiger's voice to obtain the correct information,

This experiment trained a relatively simple neural network in three methods: REINFORCE, separate method, and end-to-end, and compared its cumulative rewards. There are two internal states in Heaven/Hell problem and ten in the Tiger problem.

\begin{figure}[tb]
  \begin{center}
    \includegraphics[width=0.80\hsize]{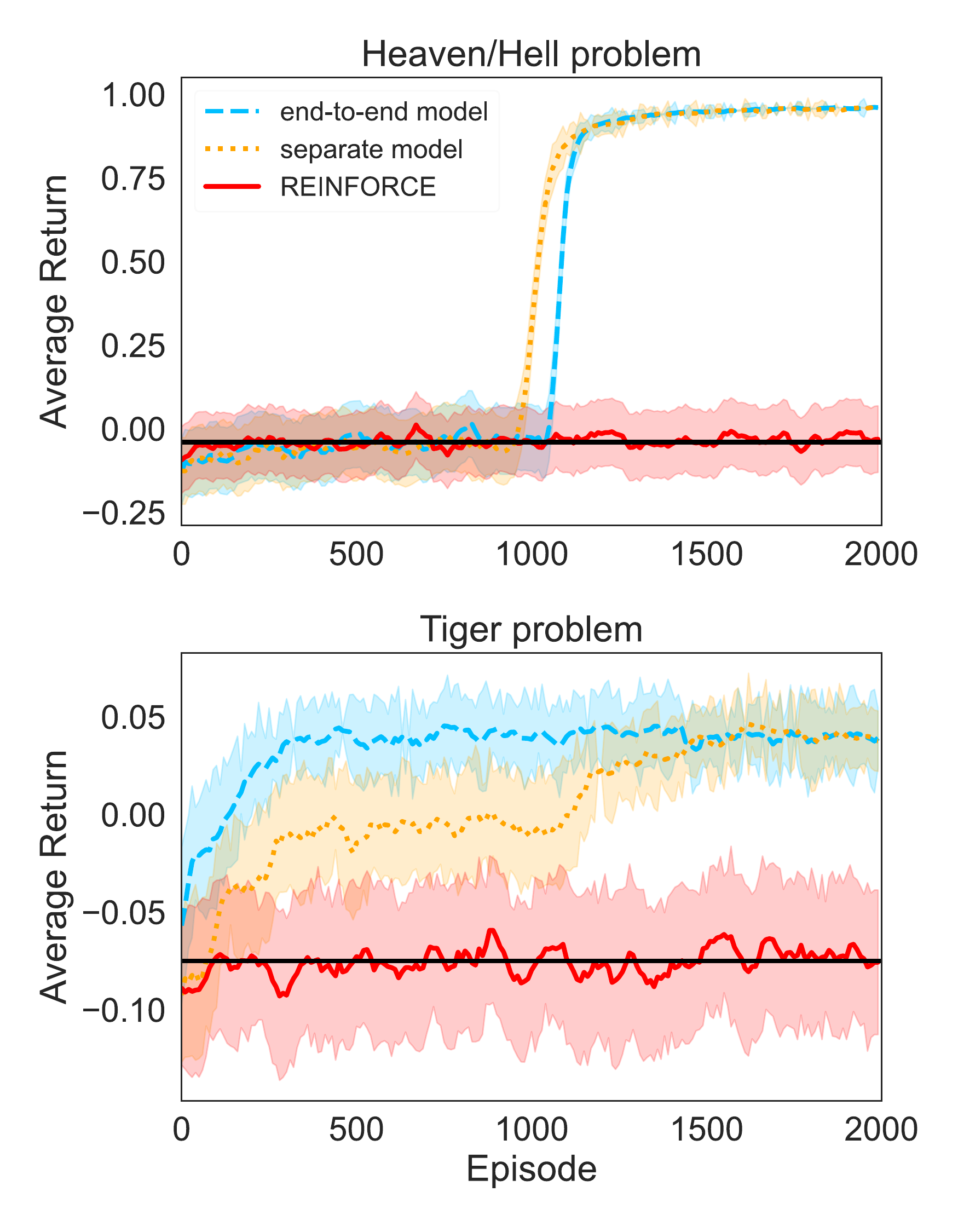}
    \caption{Training curves on Heaven/Hell problem and Tiger problem. The black horizontal line is the chance level for each problem.
    }
    \label{fig:learning_curve}
  \end{center}
\end{figure}

\begin{figure}[htbp]
  \begin{center}
    \includegraphics[width=1.\hsize]{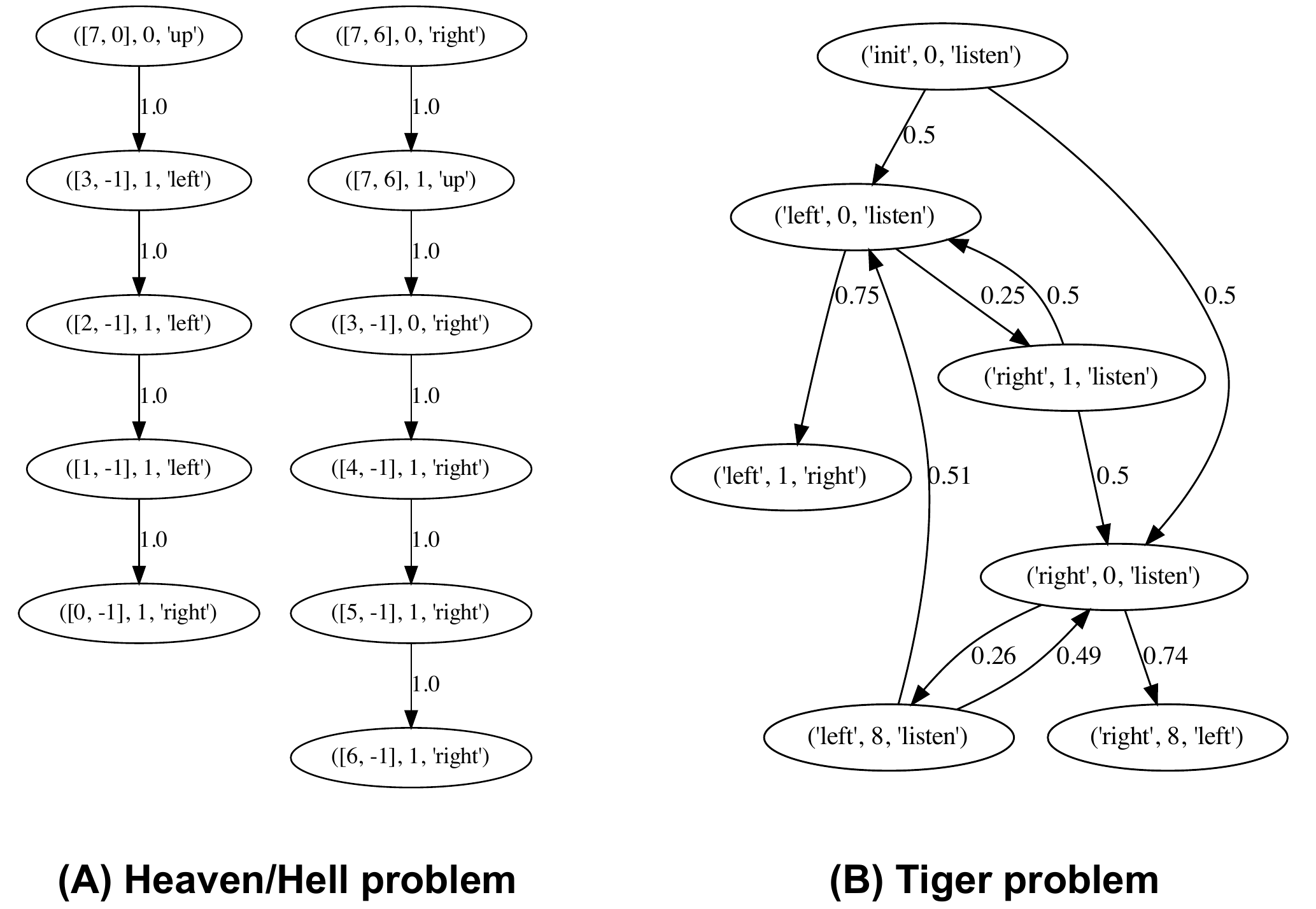}
  \caption{Transition graph over the internal states and observation pairs. In both graphs, the tuple of observation, internal state, and action is on nodes.}
  \label{fig:transition_graph}
  \end{center}
\end{figure}

\subsection{Learning Results and Visualization}
Figure~\ref{fig:learning_curve} shows the episodic average return during the training.
We trained the model in 100 environments simultaneously. The solid line corresponds to the mean and shaded range to the standard error over the 100 trials. 
In both the Heaven/Hell and Tiger problems, the separate method and the proposed method outperformed REINFORCE, which only achieved the chance level. This result suggests it is possible to solve simple POMDP problems in an end-to-end manner using the policy gradient method.

In visualization, we first collected 10000 rollouts using a trained model (we chose the action with maximum probability in the rollouts). 
Then we calculated a transition matrix of observation and internal state pairs and dropped the unreachable pairs.
Finally, the transition graphs were created (shown in Figure~\ref{fig:transition_graph}). We used a deterministic policy to create the graph; we can add the action to each node. When a policy to create rollouts is stochastic, a table of policy distribution per each pair of observations and internal states helps us analyze the agent's behavior.

As for the Heaven/Hell problem, we can say that the internal states succeeded in distinguishing the position of heaven and storing the information until the branching point and that the agent heads to the correct destination accordingly. We can see a redundant action in the case of the right heaven; the agent stays at the start point once before going up. From this observation, we can interpret that the trained policy is not optimal.

In the Tiger problem, the graph tells us that the agent opens the door opposite the door through which it has listened to the tiger's voice two or more times in a row.

As illustrated above, it is possible to interpret what the internal states represent as well as the actions over internal states and observations by seeing the transition graph, which leads to the understanding of the behavior of the trained agent. 

\section{Related Work}
RL in POMDPs requires compensation for lacking information. Depending on the approach for this challenge, the methods are classified into two categories: memory-based methods and methods using belief states.

Methods using recurrent neural networks (RNN), which compress the history into the hidden states as memory, have been proposed~\cite{hausknecht2015deep,zhu2017improving}. The dimension of the hidden state space is generally smaller than the dimension of the observation space, which is more efficient than preparing an explicit memory. On the other hand, since the hidden space is usually continuous, it takes work to interpret hidden variables and their dynamics.

A belief state is the posterior distribution of the true state calculated from history as evidence. Many works have adopted the encoder-decoder architecture for estimating belief states~\cite{igl2018deep,lee2019stochastic}.

In recent years, there have been works related to explainable AI in RL~\cite{DBLP:conf/aaai/Madumal0SV20,puiutta2020explainable}.
In this context, there are two main approaches: the model-intrinsic and model-agnostic approaches. The former approach aims to create an agent that can produce explainable outputs through training by applying some technique to the model~\cite{shu2017hierarchical}. On the other hand, the latter approach tries to obtain another model, a posteriori, that can explain the output of the learned agent~\cite{liu2018toward,amir2018highlights}. Our approach falls into the model-intrinsic approach in that we optimize the internal states through training.

\section{Discussion and Conclusion}
This study demonstrated that the end-to-end policy gradient method could optimize policy with discrete internal states in POMDPs. We also showed that visualization of the discrete internal states of an agent after training improves the interpretability of the agent's behavior. In this study, the assumption that the internal state follows the categorical distribution limits its application to relatively simple tasks. If the hidden state space is continuous or in high dimension, it becomes difficult to estimate hidden states and explain the estimated hidden states in a human-interpretable way. Future work is to train an agent in our end-to-end algorithm and visualize the estimated internal states in such a situation.

\bibliographystyle{named}
\bibliography{end-to-end}
\end{document}